                                                          \documentclass[a4paper, 10pt, conference]{ieeeconf}      % Use this line for a4
\usepackage{FG2020}

\FGfinalcopy % *** Uncomment this line for the final submission

\IEEEoverridecommandlockouts                              % This command is only
\usepackage{graphics} % for pdf, bitmapped graphics files
\usepackage{epsfig} % for postscript graphics files
\usepackage{mathptmx} % assumes new font selection scheme installed
\usepackage{float}

\def\FGPaperID{143} % *** Enter the FG2020 Paper ID here

\title{\LARGE \bf
Deformation Flow Based Two-Stream Network for Lip Reading
}

\author{\parbox{16cm}{\centering
    {\large Jingyun Xiao$^{1,2}$, Shuang Yang$^1$, Yuanhang Zhang$^{1,2}$, Shiguang Shan$^{1,2}$, Xilin Chen$^{1,2}$ }\\
    {\normalsize
    $^1$ Key Laboratory of Intelligent Information Processing of Chinese Academy of Sciences (CAS), Institute of Computing Technology, CAS, Beijing 100190, China\\
    $^2$ University of Chinese Academy of Sciences, Beijing 100049, China
    }
    }
    \thanks{This work was done by Jingyun Xiao during his internship at the Institute of Computing Technology, Chinese Academy of Sciences.}% <-this % stops a space
}

\begin{document}

\ifFGfinal
\thispagestyle{empty}
\pagestyle{empty}
\else
\author{Anonymous FG2020 submission\\ Paper ID \FGPaperID \\}
\pagestyle{plain}
\fi
\maketitle

%%%%%%%%%%%%%%%%%%%%%%%%%%%%%%%%%%%%%%%%%%%%%%%%%%%%%%%%%%%%%%%%%%%%%%%%%%%%%%%%
\begin{abstract}

   Lip reading is the task of recognizing the speech content by analyzing movements in the lip region when people are speaking. 
   Observing on the continuity in adjacent frames in the speaking process, and the consistency of the motion patterns among different speakers when they pronounce the same phoneme, we model the lip movements in the speaking process as a sequence of apparent deformations in the lip region. 
   Specifically, we introduce a Deformation Flow Network (DFN) to learn the deformation flow between adjacent frames, which directly captures the motion information within the lip region. The learned deformation flow is then combined with the original grayscale frames with a two-stream network to perform lip reading. Different from previous two-stream networks, we make the two streams learn from each other in the learning process by introducing a bidirectional knowledge distillation loss to train the two branches jointly. Owing to the complementary cues provided by different branches, the two-stream network shows a substantial improvement over using either single branch.
   A thorough experimental evaluation on two large-scale lip reading benchmarks is presented with detailed analysis. The results accord with our motivation, and show that our method achieves state-of-the-art or comparable performance on these two challenging datasets. 

\end{abstract}

%%%%%%%%%%%%%%%%%%%%%%%%%%%%%%%%%%%%%%%%%%%%%%%%%%%%%%%%%%%%%%%%%%%%%%%%%%%%%%%%
\begin{figure*}
   \centering
   \includegraphics[width=\linewidth]{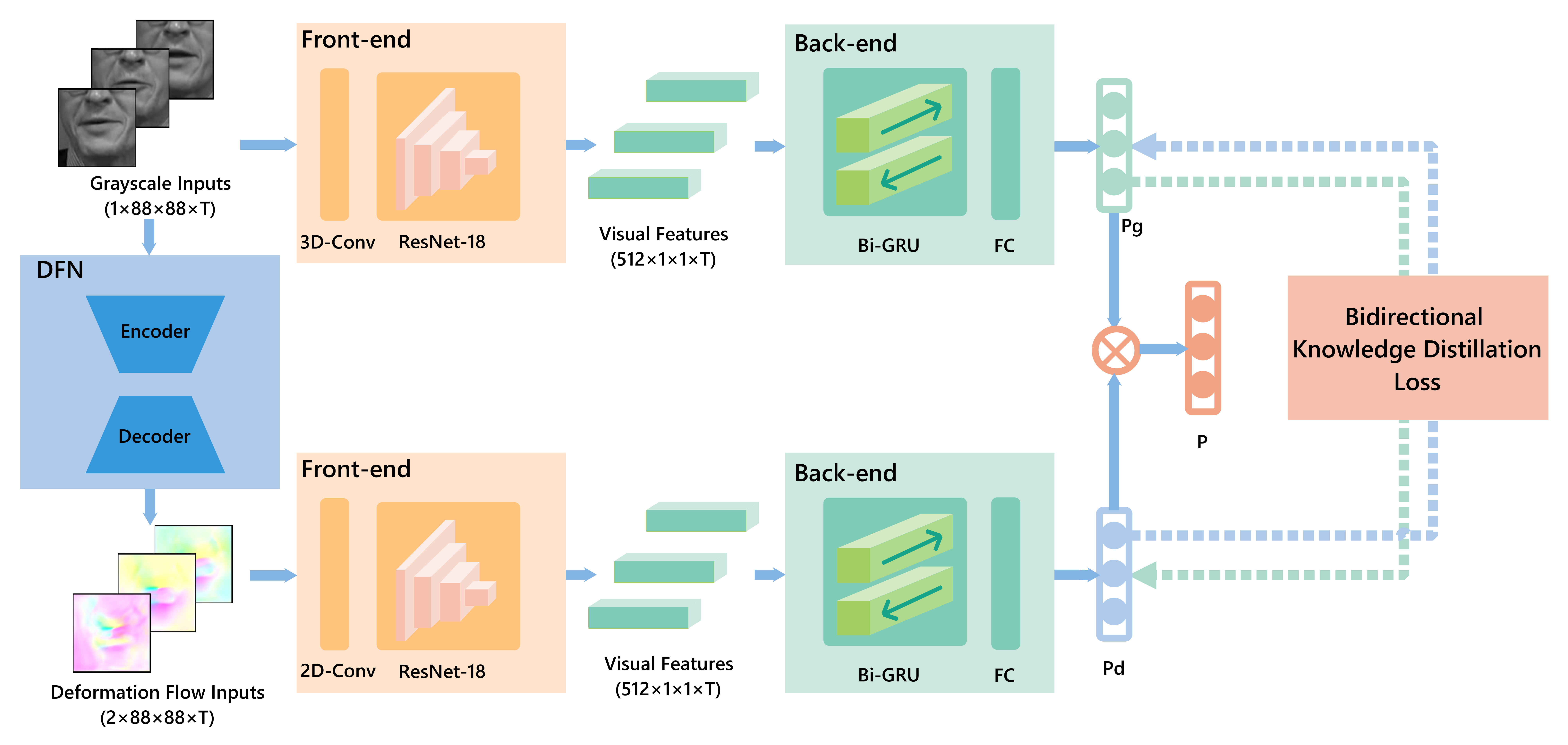}
   \caption{The overview of Deformation Flow Based Two-stream Network. Given an input video, we first feed it to the Deformation Flow Network to generate the deformation flow.  Then the raw video and the deformation flow are fed into the two branches separately. Each branch predicts the probability of each word class independently. At test time, we fuse the results of each branch to improve classification performance. During training, we propose a bidirectional knowledge distillation loss to enable the two branches exchange learned knowledge.}
   \label{fig:overview}
   \vspace{-0.4cm}
   \end{figure*}

    %Given an input video (i.e., cropped grayscale image sequence of the lip region in this paper), we first feed it into the Deformation Flow Network to generate the deformation field of every pair of adjacent frames. We get a sequence of deformation fields, which is the deformation flow of the face in the video.  Then the grayscale video and the deformation flow are fed into the two branches separately. Each branch predicts the probability of each word class independently. At the testing stage, we fuse the results of each branch and make better classification of the input video. At the training stage, we propose a bidirectional knowledge distillation loss to make the two branches exchange the learned knowledge. 

 %%%%%%%%%%%%%%            INTRODUCTION           %%%%%%%%%%%%%%%%%%%%%%%%%%%%%%%%%%%%%%%%%%%%%%%%%%%%%%%%%%%%%%%%%%
 \section{INTRODUCTION}
 %???????
 Visual speech recognition, also known as lip reading, is the task of decoding speech content based on the visual cues of a speaker's lip motion. Lip reading is a developing topic that has received growing attention in recent years. It has broad application prospects in hearing aids, special education for hearing impaired people, complementing acoustic speech recognition in noisy environments, new human-machine interaction methods, among many other potential applications.
 
 The field of video understanding has progressed significantly in recent years. However, lip reading, a special task of video understanding, remains a challenging task. Different from coarse-grained video analysis tasks, such as action detection and action recognition, lip reading is a fine-grained video analysis task, and requires subtle spatial information in the lip region as well as continuous and discriminative temporal information of lip motion. While humans outperform machines in action recognition, machines have already exceeded humans in lip reading. This is partly because the visual details and lip motions are too subtle for humans to capture and analyze, while machines have an innate advantage in this respect.
 
 Recent lip reading methods are based on deep learning and often conducted in end-to-end fashion. 
 Although promising performance has been achieved by these methods, there are still several issues that demand more consideration. 
 First, most existing lip reading methods extract frame-wise features and then model the temporal relationships with RNNs, with less consideration of the innate spatiotemporal correlation of adjacent frames. Second, one main difference between lip reading and other video analysis tasks is that the input video is focused on the face, and usually a crop of the lip region. It sets higher demands on the discriminative power of subtle facial information in the videos.
 
 In this paper, we propose a Deformation Flow Network (DFN) to generate the deformation flow of the face in a video. It is trained in a completely self-supervised manner, with no need for labeled data. The deformation flow is a sequence of deformation fields. A deformation field is a mapping of the correlated pixels from the source frame to the target frame, which directly represents the motion information from the source frame to the target frame. By computing the deformation field between each pair of adjacent frames, we can capture and represent the motion of the face in the video.
 
 For effective lip reading, we use both the computed deformation flow and the raw videos as the input to a two-stream network. The two branches predict the probabilities of each word class independently. To make the two branches exchange information during training, we adopt knowledge distillation, and utilize a bidirectional knowledge distillation loss to help the two branches learn from each other's predictions during training. At test time, we fuse predictions from both branches to make the final prediction. We observe that a simple average of the predictions produces more accurate predictions, compared with results of using either single branch. It suggests that the two sources of input, the raw video and the deformation flows, provide complementary cues for the lip reading task. 
 
 Our contributions are threefold: (a) we propose a Deformation Flow Network (DFN) to generate deformation flows that can capture the motion information of the faces, which is trained in a self-supervised manner;
 (b) we use the deformation flows and the raw videos as the inputs to a two-stream network, which provide complementary cues for lip reading, and utilize a bidirectional knowledge distillation loss to train the two branches jointly;
 (c) we conduct extensive experiments on LRW \cite{Chung2016LipRI} and LRW-1000 \cite{yang2019lrw}, demonstrating the effectiveness of our methods. 
 %???We achieve state-of-the-art performance on both datasets.

 %%%%%%%%%%%%%%%%           Related Works             %%%%%%%%%%%%%%%%%%%%%%%%%%%%%%%%%%%%%%%%%%%%%%%%%
 \section{Related Works}
 \label{sec:related_works}
 
 % ???????????????????lip reading ??

 In this section, we briefly review previous works on deep learning methods for lip reading, as well as self-supervised methods for facial deformation modeling.
 
 \subsection{Deep Learning Methods for Lip Reading}
 
 With the rapid development of deep learning in recent years, some works have begun to apply deep learning methods to lip reading and obtained considerable improvements over traditional methods using hand-engineered features.
 Noda \cite{Noda2014LipreadingUC} was the first to employ a convolutional neural network (CNN) to extract the features for lip reading for the first time. Wand et al. \cite{Wand2016LipreadingWL} used Long Short-Term Memory (LSTM) to replace the traditional classifier for lip reading, and achieved considerable improvement. 
 In 2016, Chung et al. \cite{Chung2016LipRI} proposed an end-to-end lip reading model and compared several strategies of processing the frames for word classification, which has founded a solid base for the subsequent progress for lip reading. Since then, more recent lip reading approaches have followed an end-to-end paradigm. Concurrently, Assael et al. \cite{Assael2016LipNetSL} proposed LipNet,
 which is the first end-to-end sentence-level lip reading model. 
 %It uses three cascaded spatiotemporal convolutional neural networks to extract spatiotemporal features and recurrent units to perform character prediction at each time-step. 
 %Notably, it employs the CTC loss \cite{graves2006connectionist}, which is widely used in audio speech recognition to deal with unaligned data.
 
 In 2017, Stafylakis et.al. \cite{Stafylakis2017CombiningRN} proposed a new word-level lip reading model that attains $83.0\%$ classification accuracy on the
 LRW dataset, which is a significant improvement over prior art. It uses a
 combination of a single 3D convolution layer, ResNet
 \cite{he2016deep}, and bidirectional LSTM networks \cite{hochreiter1997long}. The proposed architecture shows a strong spatiotemporal modeling power, successfully copes with many in-the-wild variations that LRW presents.
 %delete
 %To promote the discriminability of the lip reading model, Chung et.al \cite{chung2017lip} introduced attention mechanism and explored the effective attentional encoder-decoder architectures for lip reading. Afouras et al. \cite{afouras2018deep} proposed a lip reading model with self-attention layers that makes further improvements over the architectures without attention mechanism.
 Inspired by the success of deep spatiotemporal convolutional networks and two-stream architectures in action recognition, Weng et al. \cite{weng2019importance} introduced deep spatiotemporal convolutional networks to lip reading. They also employ optical flow and two-stream networks. However, the optical flow is hard to obtain and it costs considerable time and storage. Moreover, most existing optical flow methods are unable to capture the fine-grained motion information of the lip region.
 
 %delete
 %In summary, the introduction of deep learning to lip reading has yielded considerable improvements over the traditional methods using hand-engineered features. To further boost performance, researchers are exploring pertinent methods to tackle specific problems in lip reading. 

 % Based on \cite{Stafylakis2017CombiningRN}, Stafylakis et al.\cite{Stafylakis2018PushingTB} conduct detailed research on the influence of front-end and back-end in lip reading models to the performance. It also researches on the use of  word boundary information (the starting time and ending time of the target word in a video). It introduces word boundary information as addition indication in the training stage and get considerable improvement in recognition accuracy.

 \subsection{Self-supervised Facial Deformation}
 
 Recently, there have been a series of works using the deformation field and warping methods for face manipulation, facial attributes learning and other face-related tasks.
 
 The Deforming Autoencoder (DAE) \cite{shu2018deforming} presents an unsupervised method to disentangle shape (in the form of a deformation field) and appearance (texture information disregarding the pose variations) of a face. The learned features are demonstrated to be effective for face manipulation, landmarks localization and emotion estimation.
 
 X2Face \cite{Wiles2018X2FaceAN} is a network that can generate face images with a target pose and expression.
 In the evaluation stage, given a source face and a driving face, the network is able to generate a new face that preserves the identity, appearance, hairstyle and other attributes of the source face, while possessing the pose and expression of the driving face. 
 In the training stage, it uses a pixel-wise L1 loss between the generated frame and the driving frame to supervise the training process. In this way, the training process of the network does not need any annotations.
 %In the training stage, it uses a source frame and a driving frame which are sampled from the same video. The generated frame is expected to be identical to the driving frame. Therefore a pixel-wise L1 loss between the generated frame and the driving frame is used to supervise the training process. In this way, the training process of the network does not need any annotations. 
 
 %In the evaluation stage, the source frame and the driving frame can be of different identities. 
 %The generated frame is expected to possess the identity, appearance, hairstyle and other attributes of the source frame while possessing the pose and expression of the driving frame. 
 % {xx}
 
 FAb-Net \cite{Koepke2018SelfsupervisedLO} has a similar architecture to X2Face. However, it aims to learn the facial attributes in a self-supervised manner. 
 %The architecture consists of an encoder and a decoder. The encoder encodes the frame into a vector, which contains rich facial attributes information. 
 %The  decoder generates a deformation field based on the facial attribute information in the vectors. The source frame is warped by the deformation field and a generated frame is obtained for computing the pixel-wise L1 loss.  
 %In this way, the network learns facial attributes in a self-supervised manner.
 %The network uses a pixel-wise L1 loss between the generated frame and the target frame  to self-supervise the training process. 
 The learned facial attributes are demonstrated to achieve results comparable to and even surpassing the features learned by supervised methods in several tasks. 
 
 Inspired by these works, we propose the Deformation Flow Network (DFN) in our work  to model the lip motion in the speaking process for lip reading, which is also trained in a self-supervised manner.
 \section{Methods}
 \label{sec:methods}
 
 In this section, we introduce our Deformation Flow Network (DFN) for generating the deformation flow, Deformation Flow Based Two-stream Network (DFTN) for word-level lip reading, and the bidirectional knowledge distillation loss for training the two-stream network jointly. 
 
 An overview of the pipeline is shown in Fig. \ref{fig:overview}. Given an input video (i.e., cropped grayscale image sequence of the lip region), we first feed it to the Deformation Flow Network to generate a series of deformation fields, one for each pair of adjacent frames. This resulting deformation field sequence is the deformation flow of the original video. 
 Next, the grayscale video and the deformation flow are fed into the two branches separately for recognition. 
 The two branches are optimized with individual classification losses, and a bidirectional knowledge distillation loss, which helps the two branches learn from each other.  %??TODO
 At test time, we fuse the results of each branch to make the final prediction for the input video.
 
 %In the following subsections, we first state the architecture and training strategy of DFN in \ref{sec:dfn}. Next, in \ref{sec:dftn} we present the structure of the two branches of DFTN in detail. Finally, the proposed bidirectional knowledge distillation loss is explained in \ref{sec:bikd}.
 % to help the two branches learn from each other during the training process, which

 \subsection{Deformation Flow Network}
 \label{sec:dfn}
 
 % \begin{figure}
 %    \centering
 %    \includegraphics[scale=0.8]{images/dfn.pdf}
 %    \caption{The source frames, target frames fed into the DFN and the generated deformation fields and output frames.  $s_1$ and $t_1$ are of the same identity and  $s_2$ and $t_2$ are of the same identity. 
 %    $o_1 = g(s_1, t_1)$, which is expected to be the same as $t_1$. $o_2 = g(s_1, t_2)$, which is expected to have the same identity with $s_1$ and  the same facial attributes (e.g.  pose and expression) with $t2$. $o_c = g(o_2, s_1)$, which is expected to be the same as $s_1$. $o_3 = g(s_2, o_2)$. If $o_2$ has the same facial attributes with $t_2$, $o_3$ is expected to be the same as $t_2$.
 %   }
 %    \label{fig:dfn}
 %    \end{figure}
 
    % \begin{figure}[thpb]
    %    \centering
    %    %\includegraphics[scale=1.0]{figurefile}
    %    \caption{Inductance of oscillation winding on amorphous
    %     magnetic core versus DC bias magnetic field}
    %    \label{figurelabel}
    % \end{figure}
    % \vspace{-0.8cm}
    \begin{figure}
       \centering
       \includegraphics[trim=35 0 0 0, width=1.05\linewidth]{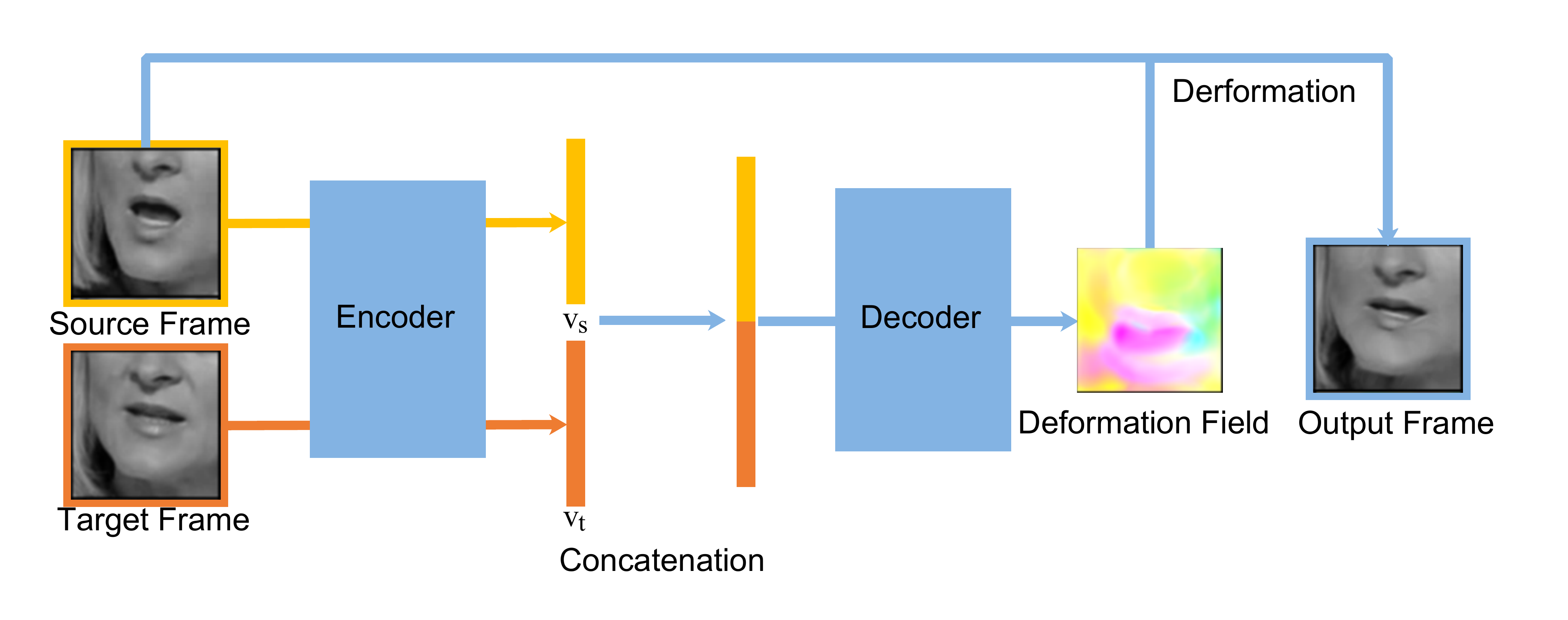}
       \caption{The architecture of DFN. It consists of an encoder and a decoder. Given a source frame and a target frame, the encoder encodes them into two feature vectors,  $v_s$ and $v_t$. The decoder takes the concatenation of $v_s$ and $v_t$ as input, and generates a deformation field. The source frame is warped by the deformation field, and generates an output frame. A pixel-wise L1 loss between the output frame and target frame can supervise the network effectively. The DFN is trained in a completely self-supervised manner.}
       \label{fig:dfn}
       \vspace{-0.4cm}
       \end{figure}
              %It consists of an encoder and a decoder. Given a source frame and a target frame, the encoder encodes them into two feature vectors,  $v_s$ and $v_t$. The decoder takes the concatenation of  $v_s$ and $v_t$ as input and generates a deformation field. The source frame is warped by the deformation field and generates an output frame. A pixel-wise L1 loss between the output frame and target frame can supervise the network effectively. The training of DFN is in a completely self-supervised manner. 

 The architecture of the Deformation Flow Network (DFN) is shown in Fig. \ref{fig:dfn}. The input to the DFN is a pair of frames (i.e.,\ a source frame and a target frame). The output of the DFN is a deformation field, which is a $2$-channel map with the same size as the input frames. The DFN consists of an encoder and a decoder. The encoder encodes the source frame $s$ and target frame $t$ into a source vector $v_s$, and a target vector  $v_t$. The decoder takes the concatenation of $v_s$ and $v_t$ as input, and generates a deformation field $d$, which predicts the relative offsets $(\delta x, \delta y)$ for each pixel location $(x, y)$ in the target frame relative to the source frame. An output frame $o$ is generated by sampling from the source frame $s$ with the offsets $(\delta x, \delta y)$ of the deformation field $d$: 
 
 \begin{equation}\label{loss_soft}
    o{(x,y)} = s{(x+\delta x, y+\delta y)}
   \end{equation}
 
 The output frame $o = D{(s, t)}$, is expected to be identical to the target frame $t$, which can be supervised by a pixel-wise L1 loss between the output frame and target frame: 
 
 \begin{equation}\label{loss_soft}
   %  $o = g(s, t)$,
   % o = g(s, t),  \mathcal{L}_1 = || o- t ||_1
    \mathcal{L}_1 = \frac{1}{n} \sum_{(x,y)} {| o(x,y) - t(x,y) |}
   % \sum_{i=0}^{n}{(x_i+y_i)}
  \end{equation}
 
\begin{figure}
  \centering
   \centering
   \includegraphics[trim=35 0 0 0, clip, width=1.05\linewidth]{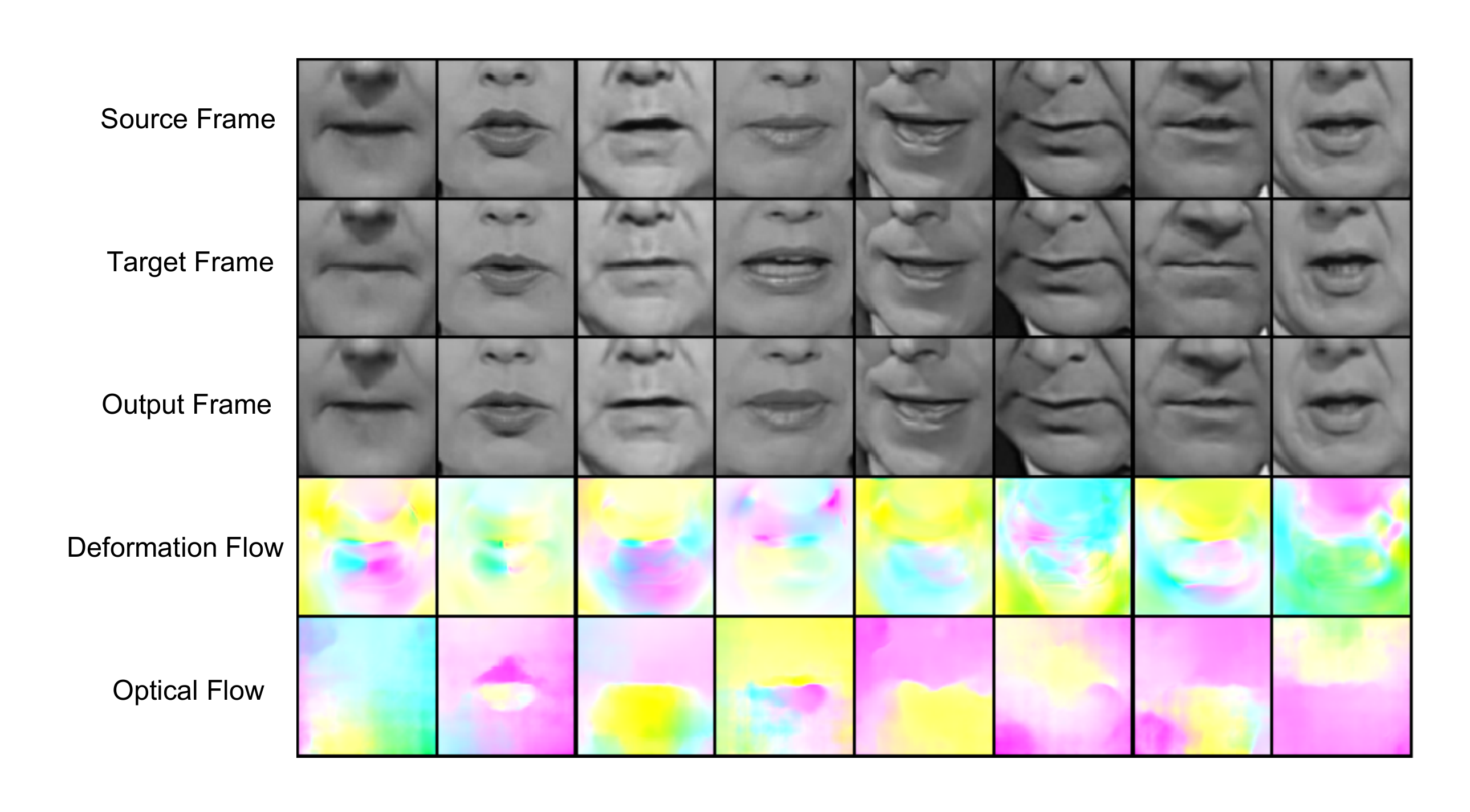}
   \caption{Examples of the source frames, target frames, output frames, deformation flow generated by DFN, and optical flow generated by PWC-Net \cite{sun2018pwc}. The color variations indicate that the deformation flow captures more details of the face than the optical flow. }
   \label{fig:sotadf}
   \vspace{-0.4cm}
   \end{figure}

 Given the above optimization target, the DFN can be trained in a completely self-supervised manner, with no need for any extra manual annotations. Examples of the source frames, target frames and output frames are shown in Fig. \ref{fig:sotadf}.
 % todo  ??DFN????????????????????
 %DFN generates deformation fields based on the facial attributes of the source frames and target frames. The deformation field predicts the offsets $(\delta x, \delta y)$ for each pixel $(x, y)$ between adjacent frames. Therefore
 It is worth noting that since the deformation field is estimated at the pixel level, it can capture very subtle variations of faces and directly represent the motion information, which means it also has great potentials in other face-related tasks beyond lip reading.
  
 %\vspace{-0.cm}
  %  \vspace{-0.cm}

   %\vspace{-0.cm}
   \begin{figure}
    \vspace{-0.2cm}
     \centering
     \includegraphics[trim=35 0 0 0, clip, width=1.05\linewidth]{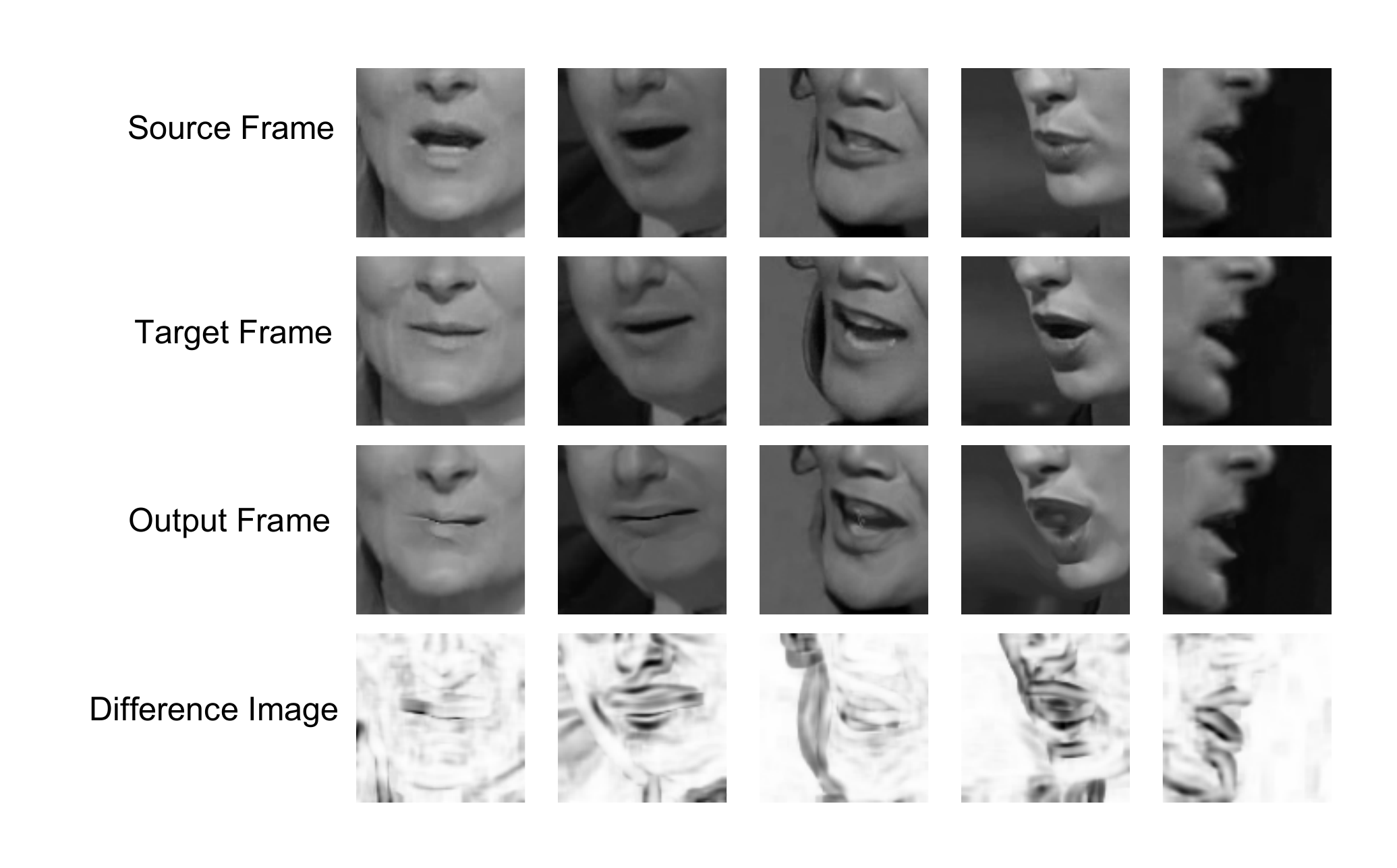}
     \caption{Examples of the difference images of the output frames and target frames.}
     \label{fig:diff}
     \vspace{-0.6cm}
     \end{figure}
    %  \vspace{-0.cm}
 
 \subsection{Deformation Flow Based Two-stream Network}
 \label{sec:dftn}
 
 In this subsection, we introduce the two branches (i.e., the grayscale branch and the deformation flow branch) of the Deformation Flow Based Two-stream Network, as well as the fusion strategy of the two branches in detail. 
 
\begin{figure}
   \centering
   \includegraphics[trim=35 0 0 0, width=\linewidth]{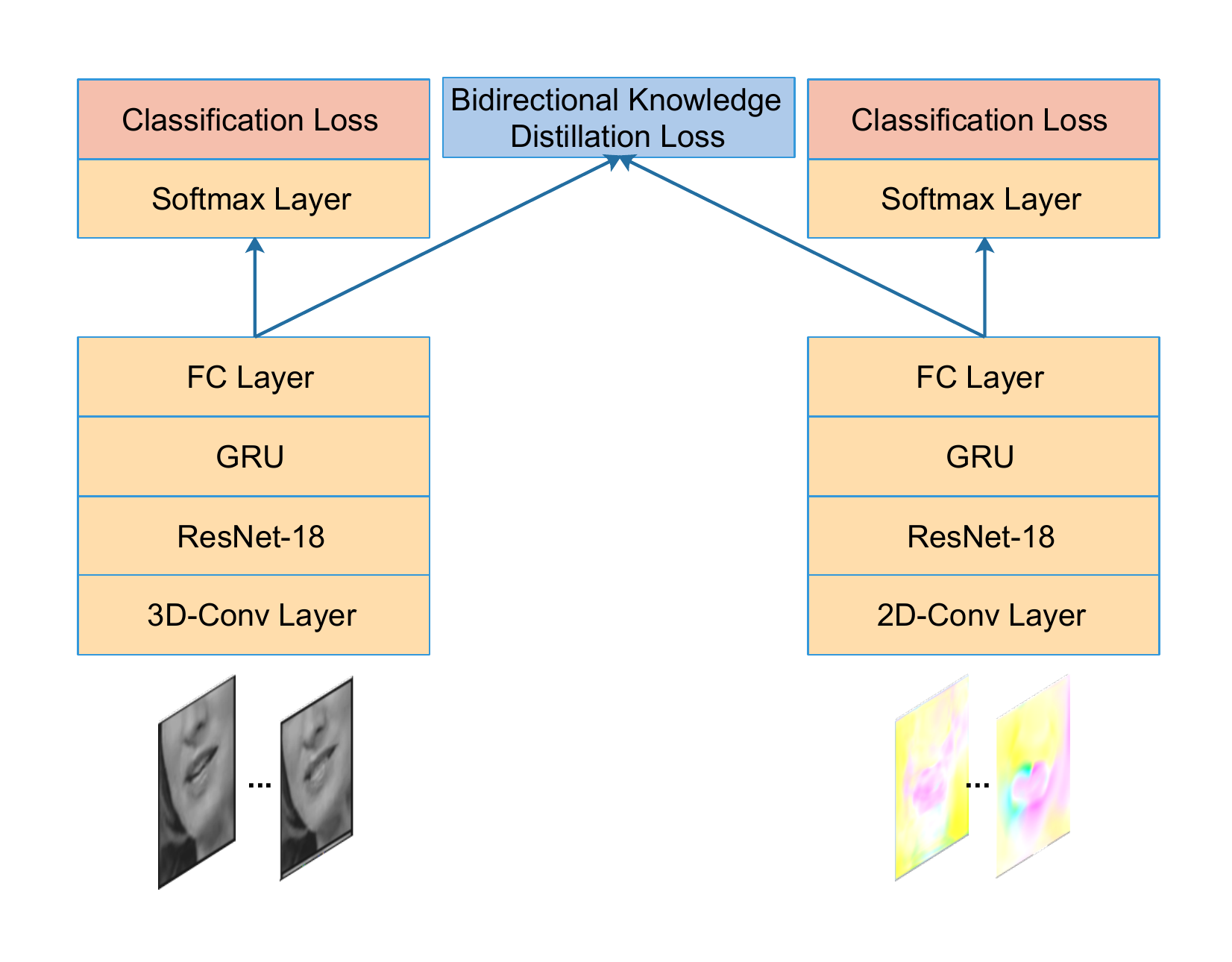}
   \caption{The architecture of the two-stream network for lip reading. The learning propcess is guided by both the classification loss and the bidirectional knowledge distillation loss.}
   \label{fig:ts}
   \vspace{-0.4cm}
   \end{figure}

 Firstly, we introduce the baseline model in this paper.
 The grayscale branch adopts the widely used architecture proposed by \cite{Stafylakis2017CombiningRN}, which is a combination of CNN and RNN, except that we use Gated Recurrent Units (GRU) \cite{chung2014empirical} instead of LSTMs. Specifically, it consists of a front-end (i.e., a single layer of 3D CNN followed by ResNet-18 \cite{he2016deep} and a back-end (i.e., a 2-layer bidirectional RNN with GRUs). The front-end extracts the visual features for each frame, and outputs a sequence of feature vectors. The back-end decodes the feature sequences, and predicts the probability of each word class. The deformation flow branch mostly mirrors the structure of the grayscale branch. The only difference is that the first layer of this branch is a 2D convolution layer, while it is a 3D convolution layer in the grayscale branch. The detailed architecture is shown in Fig. \ref{fig:ts}.
 
 Massive amount of works on two-stream networks have explored methods to fuse the two branches. In this work, we experimented with different fusion strategies, and the results with different fusion strategies are presented in \ref{subsec:exp_dfn}. 
 Among all the strategies, we find that fusing the output probabilities of the two branches gives the best performance. % In this way, the objective loss of the two-stream network is the sum of the separate classification losses of the two branches.
 % Empirically, we found that using the product of the probabilities predicted by the two branches rather than the average makes better predictions.
 
 However, the problem with fusing the predicted probabilities from individual branches is that the two branches are optimized separately, and lack interaction in the training stage. We wish to design a method that can help the two branches exchange the knowledge they learned during the training process. Therefore, we propose the bidirectional knowledge distillation loss.
 
 %\vspace{-0.cm}

 \subsection{Bidirectional Knowledge Distillation Loss}
 \label{sec:bikd}
 In this subsection, we introduce the bidirectional knowledge distillation loss as an additional supervision for training the two branches jointly.
 
 Fusion strategies for the two-stream architecture have been widely explored in the field of action recognition. Here, we adopt the method of knowledge distillation.
 The two branches are able to make word-level classification as two independent models respectively. The outputs of the fully-connected (FC) layers of the grayscale branch and the deformation branch are denoted as $\mathbf{z}_g$ and $\mathbf{z}_d$ respectively. We then obtain the predicted probability distribution over all classes, $q_g$ and $q_d$ as: 
 
 \begin{equation}\label{eqn:prob}
    q^{(i)} = \frac{\exp (z^{(i)}/T)}{\sum_{j}{\exp(z^{(j)}/T)}}, \\
 \end{equation}
 where $T$ is a parameter known as \textit{temperature}. $T$ is usually set to $1$ for classification tasks, and the equation becomes the softmax function. In knowledge distillation, a large $T$ makes the probability distribution $q$ ``softer", which is easier for a student network to learn than a one hot vector corresponding to the ground truth. In our work, we set $T$ to $20$. The knowledge distillation loss is defined as:
 \begin{equation}\label{loss_soft}
    L_{KD}(q_t, q_s) = -  \sum_{i=1}^N{q_t^{(i)}log q_s^{(i)}} ,
    \end{equation}
    where $q_t$ and $q_s$ denotes the soft probability distributions of the teacher network and student network, respectively, and $N$ denotes the number of classes. 
    
    Since we expect the two branches to learn from each other, we adopt a bidirectional knowledge distillation loss:
    \begin{equation}\label{loss_soft}
       L_{BiKD}(q_g, q_d) = L_{KD}(q_g, q_d) + L_{KD}(q_d, q_g) 
       \end{equation}
 Therefore the final objective function of the two-stream network is:
 \begin{equation}\label{loss}
    L = L_{CE}(z_g, y)+ L_{CE}(z_d, y) + \lambda L_{BiKD}(q_g, q_d) ,
    \end{equation}
 where $L_{CE}$ represents the standard cross-entropy loss for classification tasks, $y$ is the one hot vector indicating the word class label of the video, and $\lambda$ is a hyper-parameter indicating the weight of $L_{BiKD}$.
 %todo ??bikd???????????????
 
 %%%%%%%%%%%%         Experiments        %%%%%%%%%%%%%%%%%%%%%%%%%%%%%%%%%%%%%%%%%%%%%%%%%%%%%%%%%%%%%%%%%%%%
 \section{Experiments}
 \label{sec:experiments}
 
 \subsection{Datasets}
 The proposed methods are evaluated on two large-scale public lip reading datasets, LRW \cite{Chung2016LipRI} and LRW-1000 \cite{yang2019lrw}. Here we give a brief overview of the two datasets.
 
 \textbf{LRW.} LRW \cite{Chung2016LipRI} is a large and challenging word-level lip reading dataset. Each sample of LRW is a video snippet of $29$ frames captured from BBC programs. The label is the corresponding word class of the video snippet. The dataset has $500$ word classes and each class has around $1000$ training samples, $50$ validation samples and $50$ testing samples. The total duration of LRW is approximately $173$ hours. The main challenges of LRW are: (a) the variability of appearance and pose of the speakers, (b) similar word classes such as "benefit" and "benefits", "allow" and "allowed", which demands strong discriminative power of models, and (c) the target words do not exist independently in the videos; rather, they are presented with surrounding context, which requires the model to focus on the correct keyframes. 
 
 \textbf{LRW-1000.} LRW-1000 \cite{yang2019lrw} is the first public large-scale Mandarin lip reading dataset. It is a naturally-distributed large-scale benchmark for lip reading in the wild which contains $1,000$ word classes with more than $700,000$ samples from more than $2,000$ individual speakers. Each class corresponds to the syllables of a Mandarin word composed of one or several Chinese characters. It is a challenging dataset, marked by the following properties: (a) it contains significant image quality variations such as lighting conditions and scale, as well as speakers' attribute variations in pose, speech rate, age, make-up and so on, (b) the frequency of each word class is imbalanced,  which is consistent with the natural case that some words occur more frequently than others in the everyday life, and (c) the samples of the same word are not limited to a constant length range to allow for modeling of different speech rates. These factors make LRW-1000 a  challenging lip reading benchmark with a large lexicon.
 
 \subsection{Implementation Details}
 
 \textbf{Data preprocessing.}  For both LRW and LRW-1000, we resize the cropped images of lip region to $96 \times 96$ as input. For LRW, we randomly crop the input to $88 \times 88$ during training and apply random horizontal flipping. For LRW-1000, we take a central $88\times 88$ crop, and do not apply random flipping.
 
 \textbf{Network architecture.} For DFN, we employ a ResNet-18 \cite{he2016deep} as the encoder, and $7$ cascaded pairs of deconvolutional layers and bilinear upsampling layers as the decoder. The encoder yields a $256$-dimensional vector for each frame. The decoder takes the concatenation of a source vector $v_s$ and a target vector $v_t$ as input, which is $512$-dimensional, and then generates a $2$-channel deformation field with the same size as the input frames. The two channels of the deformation field  denote  the offsets along the $x$ and $y$ axis at each pixel location. 
 
 For the lip reading model, as mentioned earlier, we employ ResNet-18 as the front-end and GRU as the back-end. More specifically, for the grayscale branch, the front-end is a single 3D convolution layer followed by a powerful ResNet-18 network which yields a $512$-dimensional vector for each frame. For the deformation branch, we use a single 2D convolution layer on top of the ResNet-18 network. As for the back-end, we use  a 2-layer bidirectional Gated Recurrent Unit (Bi-GRU) RNN with 1024 hidden units to process the sequence of the $512$-dimensional vectors, each vector extracted from a frame. 
 
 \textbf{Training strategies.} We use the three-stage training strategy proposed in \cite{Stafylakis2017CombiningRN}. We use the Adam optimizer with default hyperparameters. For LRW, the learning rate is initialized to $0.0001$ and reduced by half every time when the validation loss stagnates, until the model reaches convergence. For LRW-1000, the learning rate is initialized to $0.001$. In all of our experiments, when the validation loss stagnates for the first time, we reduce  the learning rate of the back-end to $10\%$ of the learning rate of the front-end. This policy works well in alleviating the overfitting problem. As for the weight of bidirectional knowledge distillation loss, we initialize it to be $100$, and reduce it by half every time when the validation loss stagnates.
 
 \subsection{Evaluation of DFN}
 \label{subsec:exp_dfn}
 We performed a thorough evaluation of DFN over several aspects on LRW \cite{Chung2016LipRI}. 
 
 Firstly, the source frames, target frames, output frames, and generated deformation fields are shown in Fig. \ref{fig:sotadf}. As can be seen, the output frame matches the target frame quite well. Visualizations of the deformation field shows clear discrimination of the lip region, which carries the motion information we wish to capture, from neighboring regions. This indicates that DFN can generate precise deformation fields, which meets our expectation of directly capturing motion in the speakers' faces, especially in the lip region. 
 
 Secondly, we also studied the reconstruction quality of the output frames qualitatively and quantitatively. As shown in Fig. \ref{fig:diff}, DFN is able to reconstruct faces of varying poses by warping the source frames. We randomly chose $2000$ pairs of target frames and output frames to evaluate the peak signal-to-noise ratio (PSNR) and structural similarity (SSIM) index. The average PSNR is $26.86$ and the SSIM index is $0.82$, which also proves the effectiveness of our method.
 %We did not investigate further on how to improve the reconstruction quality because it is not the main concern of this paper.
 %In this work, we do not investigate further on how to improve the reconstruction quality. Because reconstructing the output frames is an intermediate step to train DFN, not a main concern of our work. We focus on the generation of deformation flow and  how to utilize it for lip reading.
 % is  SSIM 0.8243616101705019, PSNR 26.864349772138855. But we don't 
 
\begin{figure}
  \vspace{-0.6cm}
    \centering
    \includegraphics[width=\linewidth]{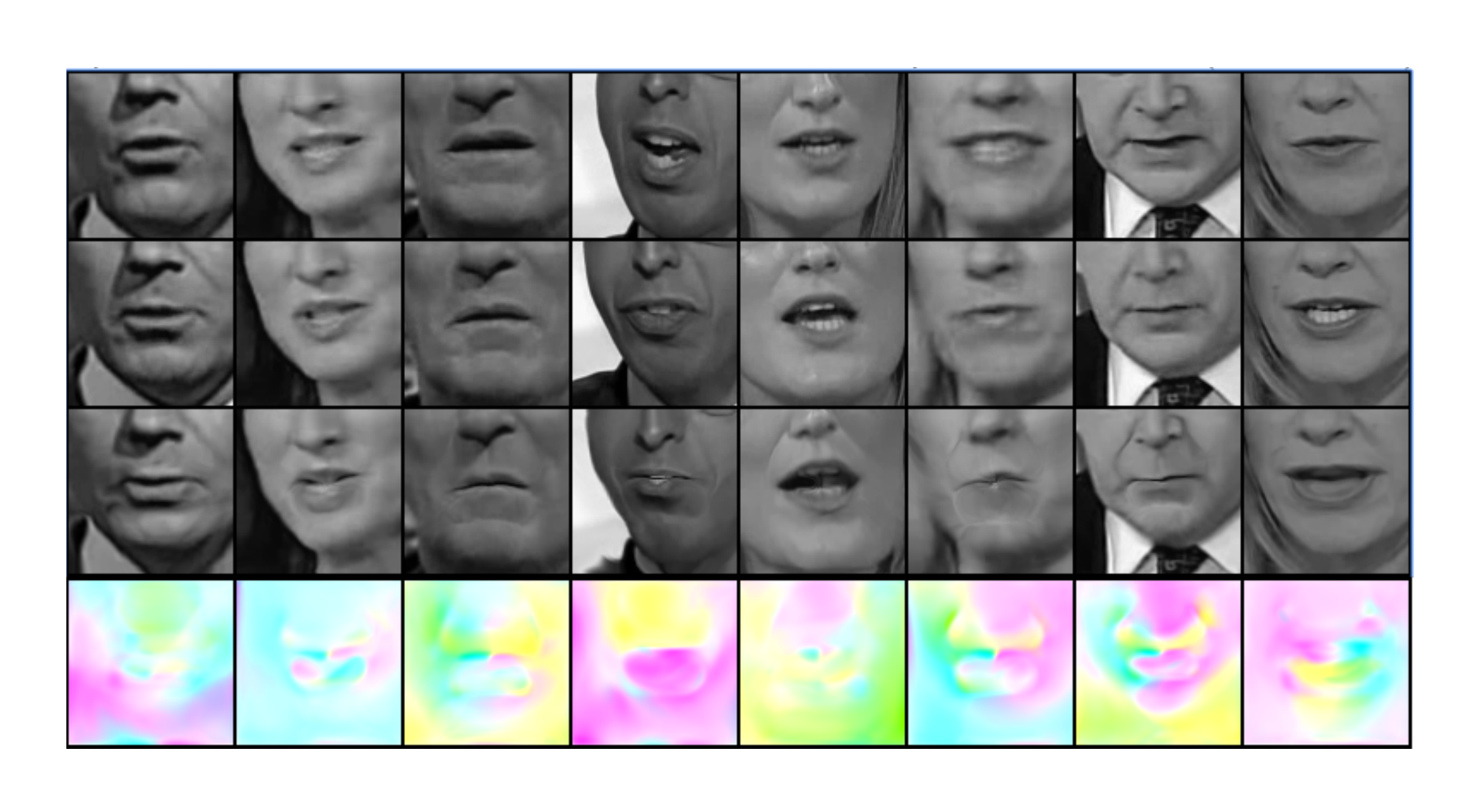}
    \caption{The output frames and deformation fields generated by DFN, where the encoder is optimized with the classification loss instead of the L1 loss. The output frames have slight differences from the target frames. According to the views in \cite{SevillaLara2017OnTI}, optical
    flow learned for action recognition in a task-specific manner differs from traditional optical flow and improves the performance of action recognition. This is also the case with the deformation flow.}
    \label{fig:encdoer}
    \vspace{-0.6cm}
    \end{figure}

 Inspired by the observations in \cite{SevillaLara2017OnTI}, we further experiment with replacing the L1 loss with classification loss to supervise the DFN. This should help the DFN learn to generate task-specific deformation flows which better suits the lip reading task. 
 %Specifically, we optimized the whole network (including DFN, front-end and back-end) of the deformation flow branch with classification loss in an end-to-end manner. We observed that when  unfreezing both the encoder and decoder of DFN, the deformation fields will be in severe distortion. However, if we freeze the decoder and only unfreeze the encoder, the deformation fields do not present severe distortion but only slight changes compared with the original deformation fields. 
 Specifically, we freeze the decoder and unfreeze the encoder of DFN when training the deformation flow branch with classification loss, after pretraining in the self-supervised manner.
 As shown in Fig. \ref{fig:encdoer}, the action of mouth opening or closing is slightly amplified in the output frames compared with the motion in the target frames. 
 %According to the views in \cite{SevillaLara2017OnTI}, training optical flow with classification loss will improve the recognition accuracy further and the learned optical flow is more suitable for action recognition compared with traditional optical flow. We found that it also holds true for the deformation flow for lip reading. 
 The classification accuracy is also improved, as shown in Table \ref{tab:dfn_result}.
 
 Finally, we compared DFN with a state-of-the-art optical flow method, PWC-Net \cite{sun2018pwc}, on the task of lip reading qualitatively and quantitatively in the following aspects. 
 
(1) We utilize the pretrained model in \cite{sun2018pwc} to generate the optical flow of the adjacent frames in the video, and use the optical flow for lip reading.  The generated optical flow and deformation flow are shown in Fig. \ref{fig:sotadf}. It shows that the deformation flow reflects more fine-grained details.

(2) We use the deformation flow generated by DFN and optical flow generated by PWC-Net as inputs to evaluate their lip reading performance on LRW respectively. The results of are presented in Table \ref{tab:dfn_result}. It indicates that our task-specific deformation flow is more effective for the lip reading task.

(3) We also compared the network complexity (i.e., floating point operations (FLOPs) and the number of params)  of DFN and PWC-Net, which is shown in Table \ref{table:complexity}. 
The result shows that the computational complexity of DFN is much lower than PWC-Net, which is one of our motivations to propose DFN.  The greatly reduced complexity makes it possible to use DFN in real-time applications.

 %We utilize the pretrained model in \cite{sun2018pwc} to generate the optical flow between the adjacent frames in the video, and use the optical flow for lip reading.  The generated optical flow and deformation flow are shown in Fig. \ref{fig:sotadf}. It shows that the deformation flow has more fine details. We then use the deformation flow generated by DFN and optical flow generated by PWC-Net as inputs to evaluate their lip reading performance on LRW. The results of are presented in Table \ref{tab:dfn_result}. It indicates that our task-specific deformation flow is more suitable for the lip reading task. We also compared the network complexity (i.e., floating point operations (FLOPs) and the number of params)  of DFN and PWC-Net, which is shown in Table \ref{table:complexity}. 
 %The result indicates that the computational complexity of DFN is much lower than PWC-Net, which is one of our motivations to propose DFN. 
 %PWC-Net and other optical flow networks deal with open scenes, while DFN focuses on faces, which is of less complexity than open scenes. Therefore we can obtain deformation flow with a light-weighted network rather than a complicated network. 
 %The greatly reduced complexity makes it possible to use DFN in real-time applications. DFN also has potential to benefit other face-related analysis tasks.
  
 %todo ??
%  \vspace{-0.4cm}
 \begin{table}
   \caption{Evaluation of different inputs on LRW.}
   \label{tab:dfn_result}
   \begin{center}
   \begin{tabular}{|l|c|}
   \hline
   Input  & Accuracy  (\%)  \\
   \hline\hline
   Grayscale & 81.91 \\
   Deformaion Flow &  77.24\\
   Deformaion Flow (optimized by classification loss) & 79.43 \\
   Optical Flow& 67.81 \\
   \hline
   \end{tabular}
   \end{center}
   \vspace{-0.4cm}
   \end{table}

  %  \vspace{-0.4cm}
   \begin{table}
     \caption{Computation expense of different networks.}
     \label{table:complexity}
     \centering
       \begin{tabular}{|c|c|c|}
         \hline
         \textbf{Network} & \textbf{GFLOPS} & \textbf{\# Params} \\
         \hline
         \hline
         DFN & $14.5$ & $7.95$M\\
         \hline
         PWC-Net & $635$ & $9.37$M\\
         \hline
         Lip Reading Model &  $18.4$ & 40.5M\\  
         \hline            
       \end{tabular}
   \end{table}
   \vspace{-0.5cm}

   \
   %     \hline
   %     \end{tabular}
   %     \end{center}
   %     \caption{The results of the baseline model with and without the feeding the fake data generated by DFN. It shows that with the fake data, the performance improves a bit.}
   %     \end{table}
   
  %  \vspace{-0.4cm}
   \begin{table}
     \caption{Evaluation of DFTN on LRW and LRW-1000.  }
      \label{tab:result_all}
      \begin{center}
      \begin{tabular}{|l|c|c|}
      \hline
      Method & LRW (\%) & LRW-1000 (\%) \\
      \hline\hline
      Grayscale branch (baseline) & 81.91 & 38.56\\
      Deformaion flow branch & 79.43 & 36.44\\
     %  Avg(sf) & \\
      Two-stream & 83.03 & 41.46\\
      Grayscale branch (with $L_{BiKD}$)&82.93 & 38.76\\
      Deformation flow branch (with $L_{BiKD}$)& 80.85 & 37.47\\
     %  Avg(sf) (with $L_{BiKD}$) & \\
      Two-stream(with $L_{BiKD}$) & \textbf{84.13 }& \textbf{41.93}\\
      \hline
      \end{tabular}
      \end{center}
      \vspace{-0.4cm}
      \end{table}
      % \vspace{-0.4cm}
 
      % \vspace{-0.4cm}
       \begin{table}
         \caption{Evaluation of different strategies on LRW.   }
          \label{tab:result_ablation}
          \begin{center}
          \begin{tabular}{|l|c|}
          \hline
          Method & Accuracy (\%) \\
          \hline\hline
          Grayscale branch & 81.91 \\
          Deformation flow branch & 79.43 \\
          Avg (FC) & 82.13\\
          Add (Res4)  & 82.52\\
          Mul (probabilities) & 83.03\\
          Mul (probabilities) (with $L_{KD(d->g)}$)&82.14\\
          Mul (probabilities) (with $L_{KD(g->d)}$)& 82.92\\
          Mul (probabilities) (with $L_{BiKD}$)& \textbf{84.13}  \\
          \hline
          \end{tabular}
          \end{center}
          \vspace{-0.4cm}
          \end{table}
          % \vspace{-0.4cm}

 \subsection{Evaluation of DFTN }
 \label{subsec:exp_single}
 
 %\vspace{-0.cm}
 \begin{figure}
  \vspace{-0.4cm}
    \centering
    \includegraphics[width=\linewidth]{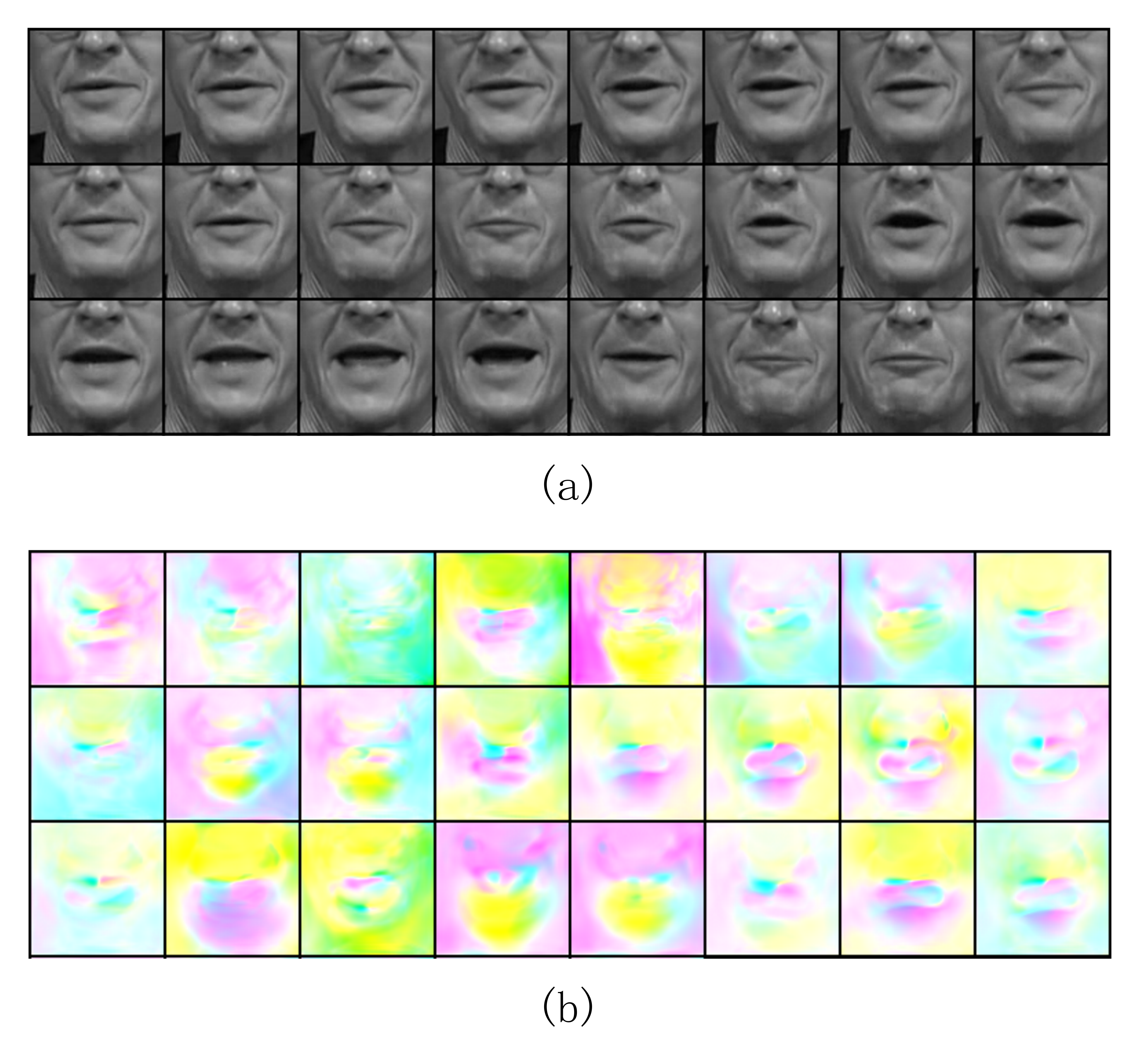}
    \caption{Examples of the inputs of the grayscale branch and the deformation flow branch.}
    \label{fig:frames_gray}
    \vspace{-0.6cm}
    \end{figure}
    % \vspace{-0.cm}
 
     In this subsection, we present the ablation studies of DFTN on LRW and LRW-1000. 
     %We evaluated the performance of each single branch as well as the two-stream network. Then we evaluated the impact of the bidirectional knowledge distillation loss on the performance of the two-stream network. Finally, we compared different fusion strategies and distilling strategies. The results demonstrate that the bidirectional knowledge distillation loss works best among all the strategies.
 
    \textbf{Evaluation of each single branch.} We pretrained the two branches (i.e., the grayscale branch and the deformation flow branch) of the two-stream networks independently. The inputs of the two branches are shown in Fig. \ref{fig:frames_gray}. The grayscale branch alone is also the baseline model in this paper. 
    %When training the deformation flow branch, we employed an end-to-end manner mentioned in \ref{subsec:exp_dfn}, updating the weights of the front-end, back-end as well as the encoder of DFN with the classification loss. 
    The results in terms of recognition accuracy on LRW and LRW-1000 are shown in Table \ref{tab:result_all}. 
 
    \textbf{Evaluation of the two-stream network.} We fused the probabilities predicted by the two branches to make the final classification of the testing samples. The results are shown in Table \ref{tab:result_all}.
    %We found that simply averaging the probabilities could yield a considerable improvement. 
    Empirically, we found using multiplicative fusion, i.e. taking the product of the probabilities results in higher recognition accuracy than additive fusion, i.e. taking the average of the probabilities of the two branches. 
    
   %  In the evaluation stage, we fuse the probability of the two branches to get the final probability distribution. Empirically, we found that using the product of the probabilities predicted by the two branches rather than the average makes better predictions.
    
    \textbf{Evaluation of the bidirectional knowledge distillation loss.} To make the two branches exchange the learned knowledge and further improve the performance of DFTN, we trained the two-stream network with the bidirectional knowledge distillation loss as an additional supervision. The results are presented in Table \ref{tab:result_all}. It is shown that the bidirectional knowledge distillation not only improves the accuracy of the joint prediction, but also improves the prediction accuracy of each branch when they work independently. 
 
    \textbf{Evaluation of different fusion strategies and distillation strategies.} To further validate the effectiveness of the bidirectional knowledge distillation loss, we conducted experiments to compare the performance of different fusion strategies and distillation strategies. We experimented with two fusion methods that fuse the intermediate features of the two branches rather than the probabilities: 
 
    \begin{enumerate}[]
     \item Average the outputs of FC layers of the two branches, feed the vector to a softmax layer to get the probability distribution, and then compute the cross-entropy loss;
     \item Adopt the fusion method in \cite{weng2019importance}, i.e. sum the outputs of the last layers of ResNet of the two branches, feed the resulting vector to the back-end to get the probability distribution, and then compute the cross-entropy loss.
     \end{enumerate}
	 Besides the above fusion strategies, we also experimented with two unidirectional knowledge distillation strategies to compare with the bidirectional knowledge distilling strategy:
 
     \begin{enumerate}[]
       \item Distill knowledge from the grayscale branch to the deformation flow branch.
       \item Distill knowledge from the deformation flow branch to the grayscale branch.
       \end{enumerate}
   
       The results are presented in Table \ref{tab:result_ablation}. It indicates that the fusion of the output probabilities performs better than the fusion of the intermediate features of the two branches (mid-fusion). Also, the bidirectional knowledge distillation outperforms unidirectional knowledge distillation with an obvious improvement. 
     
   %TODO
     %\textbf{Discussion of bidirectional knowledge distillation.} Traditional ensemble methods usually average the results of several models to approach the distribution of the datasets, which is the target distribution. Knowledge distillation is another method to achieve this goal. Instead of keeping the predicted distribution of each model unchanged, knowledge distillation motivates each model to approach the target distribution by learning from each other. And after combining the models, the prediction gets closer to the target distribution. With bidirectional distillation, we are pushing the two modalities toward the right direction.
     
     %In fact, knowledge distillation is our first trial to fuse the knowledge learned from different modalities. Our future work is to explore how to distill knowledge from different branches into a single branch, thus tackling the problem of high computational cost of two-stream architectures.
     
       % The extensive experiments on LRW and LRW-1000 demonstrate that our methods is effective.
      %  \vspace{-0.4cm}
    \begin{table}
     \caption{Comparison with other methods on LRW.  }
       \label{tab:comp_LRW}
       \begin{center}
       \begin{tabular}{|l|c|}
       \hline
       Method & Accuracy (\%) \\
       \hline\hline
      Chung16 \cite{Chung2016LipRI} & 61.10 \\
       Chung17 \cite{chung2017lip} & 76.20 \\
       Stafylakis17 \cite{Stafylakis2017CombiningRN} & 83.00 \\
       Stafylakis17 \cite{Stafylakis2017CombiningRN} (reproduced) & 77.80 \\
       % Model in \cite{Stafylakis2017CombiningRN} (with some modification) & 81.91 \\
       Weng19 \cite{weng2019importance} & 84.07 \\
     %  Model in \cite{weng2019importance} & 84.07 \\         
       DFTN & \textbf{84.13} \\
       \hline
       \end{tabular}
       \end{center}
       \vspace{-0.4cm}
       \end{table}
      %  

      %  \vspace{-0.4cm}
       \begin{table}
       \caption{Comparison with other methods on LRW-1000. }
       \label{tab:comp_lrw-1000}
       \begin{center}
       \begin{tabular}{|l|c|}
       \hline
       Method & Accuracy (\%) \\
       \hline\hline
       Yang19 \cite{yang2019lrw} &  38.19\\
       Wang19 \cite{wang2019multi} &  36.91\\
       DFTN & \textbf{41.93} \\
       \hline
       \end{tabular}
       \end{center}
       \vspace{-0.4cm}
       \end{table}
      %  \vspace{-0.4cm}
 %todo ??
 
   % \textbf{Evaluation of DFTN on LRW-1000} We also validated the effectiveness of DFTN on LRW-1000, the first Madarin language lip reading dataset. We trained the two branches respectively and jointly as we did in \ref{subsec:exp_single} on LRW-1000. Note that in all of our experiments, we loaded the model pretrained on LRW and fintune them on LRW-1000. The results are shown in Table \ref{tab:result_lrw-1000}. We also compare our method with the baseline method in \cite{yang2019lrw} on LRW-1000. Our method achieves $1.7 \% $ relative improvement. It shows that our method is effective on different large-scale datasets.
 
 \subsection{Comparison with State-of-the-Art}
 
 \textbf{Comparison with other methods on LRW.} We compared our method with other word-level lip reading methods \cite{Chung2016LipRI, Stafylakis2017CombiningRN, weng2019importance} on LRW. 
 %We reimplemented the model in \cite{Stafylakis2017CombiningRN} with PyTorch and get the  accuracy equal of $77.80\%$, lower than the result reported in \cite{Stafylakis2017CombiningRN}. We then made some modifications to the model in \cite{Stafylakis2017CombiningRN} to be our baseline model. 
 The results are presented in Table \ref{tab:comp_LRW}.  The model in \cite{weng2019importance} employs deep 3D CNNs and optical flow based two-stream networks, which achieved the existing state-of-the-art performance. Our method outperforms it, and establishes the  new state-of-the-art performance. 
 
 \textbf{Comparison with other methods on LRW-1000.} We compared our method with other word-level lip reading methods \cite{yang2019lrw, wang2019multi} on LRW-1000. The results are presented in Table \ref{tab:comp_lrw-1000}. Our method shows a considerable improvement over all previous methods on LRW-1000 and achieves state-of-the-art performance.

 % \begin{table}
 %   \caption{Performance evaluation on LRW-1000. }
 %   \label{tab:result_lrw-1000}
 %   \begin{center}
 %   \begin{tabular}{|l|c|}
 %   \hline
 %   Method & Accuracy \\
 %   \hline\hline
 %   Baseline (RGB branch) & 38.564448 \\
 %   DF branch &  36.437529\\
 %   $L_d+L_g+L_{BiKD}$&   \\
 %   RGB branch (with $L_{BiKD}$)&\\
 %   DF branch (with $L_{BiKD}$)& \\
 %   \hline
 %   \end{tabular}
 %   \end{center}
 %   \end{table}

 % As for the weight of bidirectional knowledge distillation loss, we initialize it to be 100, and reduce it by half every time the validation loss fluctuates.
 
 % First, we try using the average of the output of the softmax layer and the outputs of the 4th layer of ResNet. Then we notice that keeping their separate classification loss works better. We then conduct ablation experiments on the impact of knowledge distilling loss by applying two unidirectional experiments and a bidirection experiments. We also compare our methods with other word-level lip reading methods.
 
 % xxxxx
 
 %%%%%%%%%%%%%%%%%%%%%%%%%%%%%%%%%%%%%%%%%%%%%%%%%%%%%%%%%%%%%%%%%%%%%%%%%%%%%%%%
 \section{Conclusion}
 \label{sec:conclusion}
 
 In this paper, we propose a Deformation Flow Network (DFN) to generate the deformation flow, a way to model the lip movements in the speaking process as a sequence of deformations over the lip region.  
 %facial motion, which provides cues complementary to the raw videos. 
 Notably, the network is lightweight and trained in a self-supervised manner. 
 To take advantages of the  complementary cues provided by the deformation flow and the raw videos, we propose a Deformation Flow Based Two-stream Network (DFTN) for word-level lip reading. Different from previous methods that fuse the features of the two branches, we employ the bidirectional knowledge distillation loss to help the two branches interact with each other, and exchange knowledge during training. 
 %Trained in this way, not only the joint performance of the two-stream network gets improved, but the performance of each single branch  gets promoted as well. 
 Finally, we compare our method with other word-level lip reading methods, and show that our method achieves state-of-the-art performance. 
 Our work makes a first attempt to introduce facial deformation to generate a new modality. It provides potential applications and possibilities for not only lip reading, but also other face analysis tasks. 
 
 \section{ACKNOWLEDGMENTS}

This work is partially supported by National Key R\&D Program of China (No. 2017YFA0700804) and National Natural Science Foundation of China (No. 61702486, 61876171).
 %This work is partially supported by National Key R&D Program of China (No. 2017YFA0700804) and National Natural Science Foundation of China (No. 61702486, 61876171). 
 %%%%%%%%%%%%%%%%%%%%%%%%%%%%%%%%%%%%%%%%%%%%%%%%%%%%%%%%%%%%%%%%%%%%%%%%%%%%%%%%
 \bibliographystyle{ieee}
 \bibliography{refs}

 \end{document}